\documentclass[sn-mathphys-num]{sn-jnl}

\usepackage{graphicx}%
\usepackage{multirow}%
\usepackage{amsmath,amssymb,amsfonts}%
\usepackage{amsthm}%
\usepackage{mathrsfs}%
\usepackage[title]{appendix}%
\usepackage{xcolor}%
\usepackage{textcomp}%
\usepackage{manyfoot}%
\usepackage{booktabs}%
\usepackage{algorithm}%
\usepackage{algorithmicx}%
\usepackage{algpseudocode}%
\usepackage{listings}%

\theoremstyle{thmstyleone}%
\theoremstyle{thmstyletwo}%

\theoremstyle{thmstylethree}%

\begin{document}

\title[Article Title]{Synthesizing Efficient Data with Diffusion Models for Person Re-Identification Pre-Training}

\author[1]{\fnm{Ke} \sur{Niu}}\email{kniu22@m.fudan.edu.cn}

\author[1]{\fnm{Haiyang} \sur{Yu}}\email{hyyu20@fudan.edu.cn}

\author[2]{\fnm{Xuelin} \sur{Qian}}\email{xuelinq92@gmail.com}

\author[1]{\fnm{Teng} \sur{Fu}}\email{fut21@m.fudan.edu.cn}

\author[1]{\fnm{Bin} \sur{Li}}\email{libin@fudan.edu.cn}

\author*[1]{\fnm{Xiangyang} \sur{Xue}}\email{xyxue@fudan.edu.cn}

\affil[1]{\orgname{Fudan University}, \city{Shanghai}, \country{China}}

\affil[2]{\orgname{Northwestern Polythechnical University}, \city{Xi'an}, \country{China}}

\abstract{Existing person re-identification (Re-ID) methods principally deploy the ImageNet-1K dataset for model initialization, which inevitably results in sub-optimal situations due to the large domain gap. One of the key challenges is that building large-scale person Re-ID datasets is time-consuming. Some previous efforts address this problem by collecting person images from the internet (\textit{e.g.}, LUPerson), but it struggles to learn from unlabeled, uncontrollable, and noisy data.
In this paper, we present a novel paradigm \textit{Diffusion-ReID} to efficiently augment and generate diverse images based on known identities without requiring any cost of data collection and annotation. 
Technically, this paradigm unfolds in two stages: generation and filtering. During the generation stage, we propose Language Prompts Enhancement (LPE) to ensure the ID consistency between the input image sequence and the generated images. In the diffusion process, we propose a Diversity Injection (DI) module to increase attribute diversity. In order to make the generated data have higher quality, we apply a Re-ID confidence threshold filter to further remove the low-quality images. Benefiting from our proposed paradigm, we first create a new large-scale person Re-ID dataset \textit{Diff-Person}, which consists of over 777K images from 5,183 identities.
Next, we build a stronger person Re-ID backbone pre-trained on our \textit{Diff-Person}.
Extensive experiments are conducted on four person Re-ID benchmarks in six widely used settings.
Compared with other pre-training and self-supervised competitors, our approach shows significant superiority. Our codes and datasets will be available at \url{https://github.com/KeNiu042/Diffusion-ReID}.}

\keywords{Person Re-identification, Diffusion Model, Pre-training, Synthetic Data. }

\maketitle

\section{Introduction}\label{sec1}

Person Re-identification (Re-ID)~\cite{wei2018person,zheng2015scalable,li2014deepreid,Wang_Yuan_Chen_Li_Zhou_2018} is a task that involves cross-camera retrieval. The goal of person Re-ID is to retrieve images of the target person from a database, which is collected from multiple non-overlapping cameras. 
However, collecting and annotating a dataset including diverse person IDs and sufficient samples requires significant time and human resources. Consequently, the challenges have become obstacles hindering further development in person Re-ID. Unlike other computer vision tasks, data collection and annotation in this task are challenging and costly due to recent concerns about data privacy and usage rights.

In the past decade, deep learning, driven by extensive annotated data, has promoted the field of person Re-ID. Existing publicly annotated datasets have certain limitations regarding image quantity, captured identities, and environments. 
MSMT17~\cite{wei2018person} is a widely-used annotated dataset containing 126K images from 4K identities. However, as other biometric recognition tasks like face recognition, person Re-ID also needs large-scale datasets. For instance, IMDb-Face~\cite{wang2018devil} encompasses approximately 1.7 million face images from 59K identities. 
Besides, most of the currently available annotated person Re-ID datasets are collected in a campus environment, which makes the dataset relatively homogeneous in context and limits the further development of generalization person Re-ID. In order to solve the dilemma of data scarce, LUPerson~\cite{fu2021unsupervised} and SYSU30K~\cite{wang2020weakly} have collected a large number of unlabeled data from various online channels, which still requires significant labor costs to filter out invalid samples.

Another solution for the shortage of person Re-ID data is mainly based on synthesizing data.
Synthetic data typically originates from traditional modeling or deep generative models, which are direct and effective methods. Traditional modeling generates real-world patterns based on prior expert knowledge through mathematical models while deep generative models are designed to learn patterns from training datasets automatically. Traditional modeling approaches like RandPerson~\cite{wang2020surpassing} and PersonX~\cite{sun2019dissecting} manually simulate numerous virtual environments and compose hand-crafted 3D person models to investigate variations in visual factors such as viewpoint, pose, illumination, and background. However, some drawbacks are inevitable, such as noticeable domain gaps from real-world data, limited data quantity, and a lack of diversity.
Previous studies on deep generative models have primarily focused on generative adversarial networks (GANs)~\cite{goodfellow2020generative}, utilizing image-image style transfer for data augmentation.
Nonetheless, prior modeling efforts have emphasized transferring existing images within fixed patterns, neglecting the introduction of novel attributes such as new clothing, carried items, actions, camera poses, and scenes.
Given the limitations of the aforementioned generative models, the prevalent approach involves utilizing backbones pre-trained on ImageNet-1K~\cite{imagenet15russakovsky} for improved model initialization. However, this strategy inevitably faces sub-optimal outcomes owing to the substantial domain gap between ImageNet-1K and person Re-ID datasets.

Overall, two requirements are indispensable to generate data for person Re-ID: ID consistency and attribute diversity. ID consistency, which means a robust correlation identity between input image sequence and synthetic data, is necessary for the convergence of person Re-ID model training. Attribute diversity aims to randomly synthesize person images with diverse attributes, such as clothing and scenes that are not presented in the input image sequence. Attribute diversity ensures that the backbone encounters numerous appearances during the pre-training phase, strengthening its generalization capabilities in the fine-tuning phase.

\begin{figure*}[t]
\centering
\includegraphics[width=0.85\linewidth]{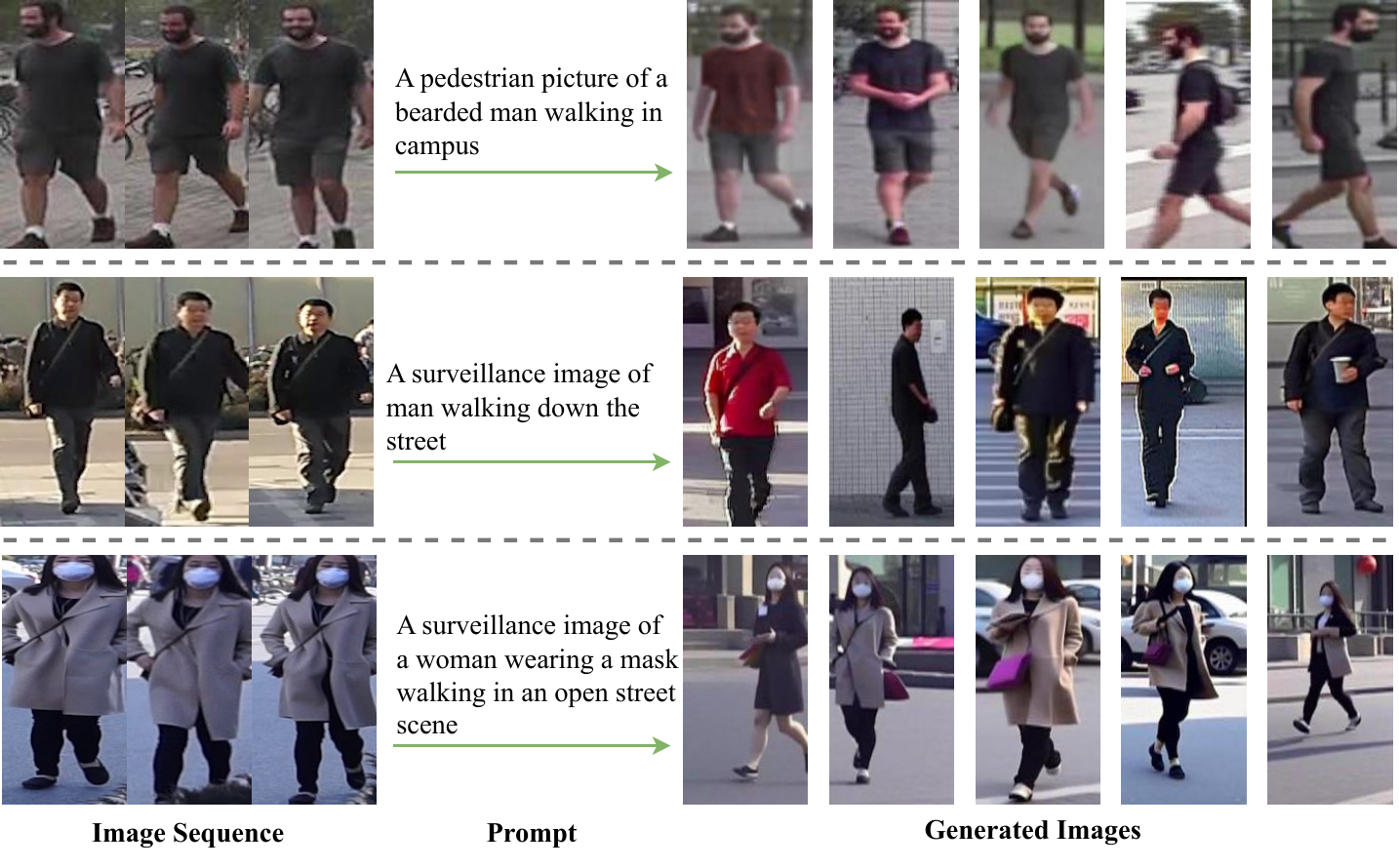}
\caption{
\textbf{Example images generated by our proposed paradigm \textit{Diffusion-ReID}.}}
\label{fig:Visualization}
\end{figure*}

In this work, we develop a paradigm termed \textit{Diffusion-ReID} for generating person images. This paradigm unfolds in two stages: the generation stage and the filtering stage. The generation stage is based on text-to-image diffusion models, trained with an image sequence with a specific ID and corresponding category-level natural language description as inputs. During the generation stage, we propose the Language Prompts Enhancement (LPE) module to ensure the ID consistency between the input image sequence and the generated images. Specifically, we utilize the image sequence captioning model to generate the prompts for the input image sequences. In LPE, we incorporate an Identity Information Representer (IIR) into the acquired prompt, facilitating the mapping of ID information at the feature level. In the diffusion process, we propose a Diversity Injection (DI) module to increase attribute diversity. We use the pre-trained diffusion model to generate an Attribute Reference Set and calculate the Fine-Grain-Specific Prior Preservation Loss with the generated images to fine-tune the diffusion model. To make the generated data have higher quality, we apply a Re-ID confidence threshold filter to further remove the low-quality images.
This versatile paradigm is easily adaptable to any person Re-ID dataset, addressing data deficiency and imbalance. We have created an annotated person Re-ID pre-training dataset, \textit{Diff-Person}, which contains over 777K images from 5,183 identities. Fig.~\ref{fig:Visualization} shows that the generated images exhibit substantial diversity in clothing, carried items, actions, camera poses, and scenes not present in the input image sequence. We pre-train a better backbone using this annotated dataset, which contains far smaller data quantity and showcases its effectiveness for various person Re-ID settings. We compare the performance of our pre-trained backbone and ImageNet-1K pre-trained backbone for six widely used settings: the supervised and unsupervised person Re-ID settings, the few-shot and the small-scale Re-ID settings, the domain adaption Re-ID setting, and the domain generalization Re-ID setting.

The contributions of this paper can be summarized as follows:
\begin{itemize}
    \item Based on diffusion models, we develop a pedestrian data generation and filtering paradigm \textit{Diffusion-ReID}, which can efficiently expand the existing annotated datasets with ID consistency and attribute diversity.
    \item We establish \textit{Diff-Person}, an annotated person Re-ID pre-training dataset composed of over 777K images from 5,183 identities. This dataset is a significant step towards scaling up the existing dataset and addressing the data deficiency and data imbalance challenges in person Re-ID.
    \item We pre-train a person Re-ID backbone based on \textit{Diff-Person}, which achieves a boost compared to the currently widely used ImageNet-1K pre-trained backbone for six widely used settings.
\end{itemize} 

\section{Related Work}\label{sec2}

\subsection{Synthetic Data for Person Re-Identification}

Most studies of person Re-ID employ supervised learning~\cite{Wang_Yuan_Chen_Li_Zhou_2018,pami21reidsurvey,he2020fastreid}. However, supervised training of Re-ID models with small-scale datasets suffers from inferior performance and poor generalization. Thus, synthetic data are often used to solve the problem of data insufficiency in the person Re-ID task. RandPerson~\cite{wang2020surpassing} developed a method for generating 3D characters with various clothes, races, and attributes and then simulated virtual environments using Unity3D. PersonX~\cite{sun2019dissecting} utilizes an open-source synthetic data engine, PersonX, to compose hand-crafted 3D person models. However, these approaches have several drawbacks: 1) They introduce noticeable discrepancies from real-world data; 2) Storing, sharing, and transferring numerous virtual environments and composing hand-crafted 3D person models can be costly; 3) The specific data source limits data quantity and diversity. Previous studies on deep generative models have primarily focused on GANs, which fall into two categories: image-image style transfer and data augmentation. For image-image style transfer methods, CamStyle~\cite{zhong2018camstyle} facilitates the conversion between any two image styles, while PTGAN~\cite{wei2018person} proposes a person transfer from the source domain to the target domain. Data augmentation-based methods commence with model training to improve generalization abilities. PNGAN~\cite{qian2018pose} employs pose normalization GAN to produce person images with uniform body poses. 
However, they suffer from some shortcomings: 1) The generation fixed patterns and can only transform existing materials; 2) These methods can only generate small-scale data according to predefined settings; 3) The quality of the generated images is affected by the original image.
Our work explores the latest text-to-image diffusion model and further captures the essential requirement of person Re-ID generation: ID consistency and attribute diversity to generate an arbitrary number of high-quality images independent of the input image. 

\subsection{Text-to-Image Diffusion Model}

Diffusion model as a likelihood-based model has recently been widely used in generative tasks, and its broad application prospects and unparalleled generative effects have attracted widespread attention. One of the popular developments is text-to-image diffusion models that synthesize fidelity images conditioned on any given text prompts. Imagen~\cite{saharia2022photorealistic} proves that pre-trained text embeddings are remarkably effective for text-to-image synthesis. DALL-E2~\cite{ramesh2022hierarchical} combines CLIP~\cite{radford2021learning} and diffusion model to improve sample fidelity at the cost of sample diversity. Stable Diffusion~\cite{rombach2022high} trains diffusion models in the latent space of pre-trained autoencoders to get high-resolution image.
VI-Diff~\cite{huang2023vi} is a method for Visible-Infrared person Re-ID to generate visually realistic images. However, as with previous generative person Re-ID work, it does not allow for the introducing of new unseen attributes. DreamBooth~\cite{ruiz2022dreambooth}, which our model is built upon, is proposed to personalize the contents in the generated results using a small set of images with the same categories. We found it very suitable for increasing the diversity of few-shot images. 
However, there are still some issues with applying this method directly to person Re-ID.

\section{Method}\label{sec4}
\begin{figure*}[t]
    \centering
    \includegraphics[width=1\textwidth]{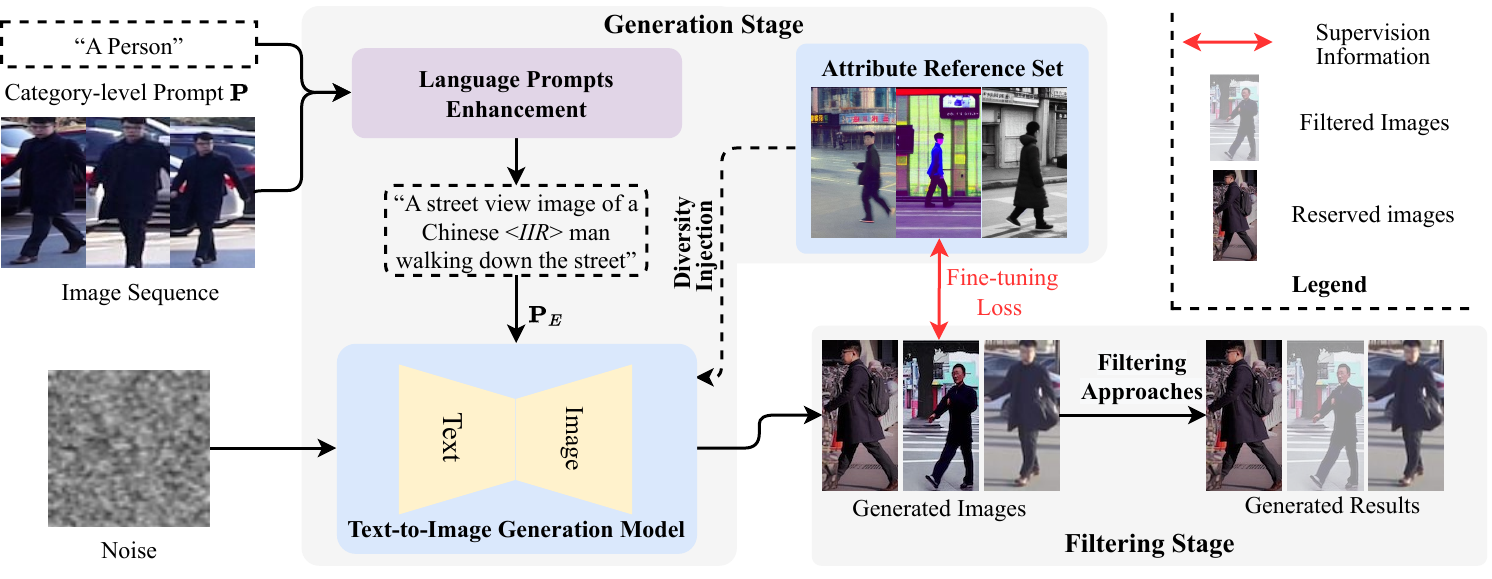}
    \caption{\textbf{The overview of our paradigm \textit{Diffusion-ReID}.} } 
    \label{fig:overview}
\end{figure*}

The proposed paradigm, illustrated in Fig.~\ref{fig:overview}, unfolds in the generation and filtering stages. The generation stage mainly consists of a text-to-image diffusion model. Different from existing diffusion models, we additionally introduce the Language Prompts Enhancement (LPE) module and the Diversity Injection (DI) module to ensure ID consistency and attribute diversity, respectively. Specifically, the LPE module takes specific ID image sequences and the category-level prompt $\mathbf{P}$ as inputs to produce the enhanced prompt $\mathbf{P}_E$ with fine-grained local details and global contextual information through a pre-trained image captioning model. In $\mathbf{P}_E$, we incorporate an Identity Information Representer $<\!\!IIR\!\!>$ to map ID information between text embedding and image embedding at the feature level. In the diffusion process, we propose a Diversity Injection (DI) module to boost attribute diversity. Concretely, we use the pre-trained diffusion model to generate an Attribute Reference Set and calculate the Fine-Grain-Specific Prior Preservation Loss with the generated images to fine-tune the diffusion model. In the filtering stage, we apply a Re-ID confidence threshold filter to remove low-quality images.

In the following, we present the generation and filtering stages in Section~\ref{4.1} and ~\ref{4.2}, respectively. Then, we introduce a novel person Re-ID pre-training dataset created by the above paradigm without any further annotation work in Section~\ref{4.3}. 

\subsection{Generation Stage}
\label{4.1}
For the generation stage, we introduce two modules, LPE and DI, to fine-tune the diffusion model. LPE aims to ensure ID consistency by enriching category-level natural language description and incorporating an IIR; DI boosts attribute diversity by fine-tuning the diffusion model with the Fine-Grain-Specific Prior Preservation Loss.

\textbf{Language Prompts Enhancement(LPE)}. Text-to-image diffusion models in latent space rely on multi-modality models, where image embeddings and text embeddings are matched in the same representation space. Utilizing a pre-trained text-to-image generation model involves constructing text prompts based on target categories to synthesize corresponding images. However, challenges arise when applying previous diffusion models to generate person images. Firstly, existing models tend to be coarse-grained, prioritizing category information over ID consistency, impacting the capture of fine details crucial for identity, such as facial features and body shape. Secondly, these pre-trained diffusion models struggle to align with the style of real-world street view camera settings, resulting in the generation of unreasonable images, including wrongly simulated pedestrian environments and inaccuracies in lighting.

\begin{figure}[t]
    \centering
    \begin{minipage}{0.49\textwidth}
        \centering
        \includegraphics[width=\linewidth]{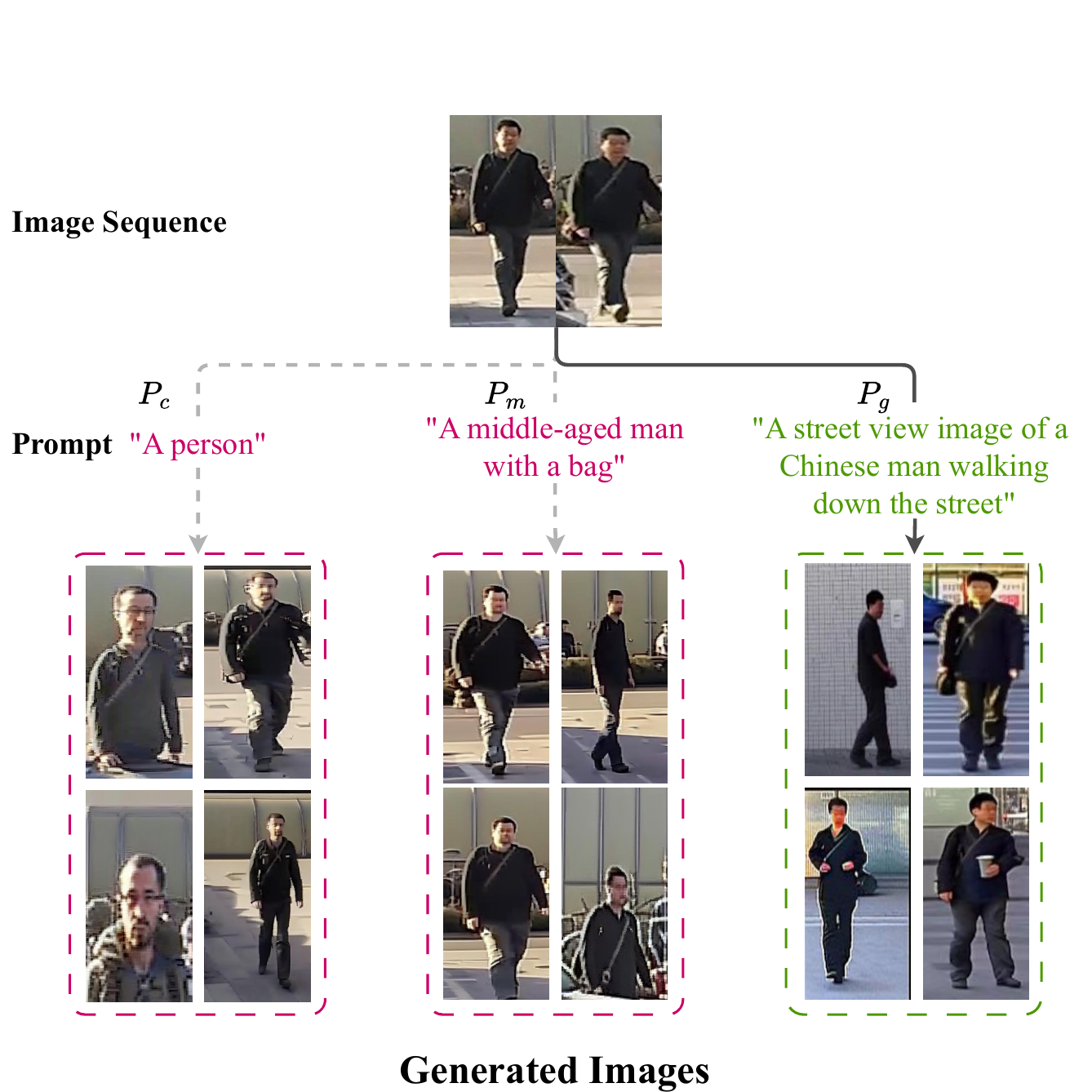}
        \caption{
 \textbf{Comparison of generated results between manual text prompt input and captioning models}.
 }
        \label{fig:lpe}
    \end{minipage}
    \begin{minipage}{0.49\textwidth}
        \centering
        \includegraphics[width=\linewidth]{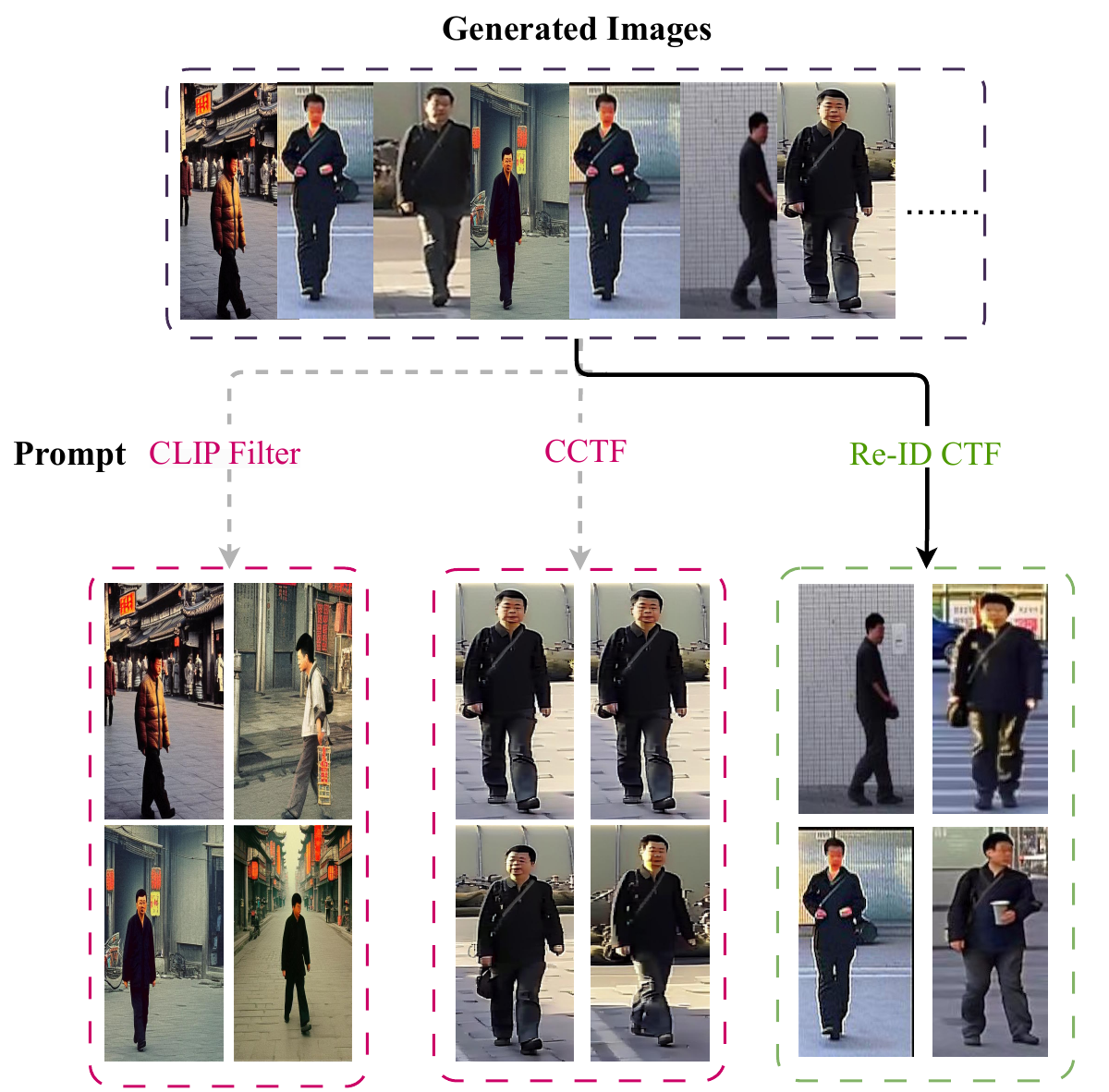}
        \caption{\textbf{Illustration of three proposed candidate filtering approaches, as well as the output visualization.}
 }
        \label{fig:filter}
    \end{minipage}
\end{figure}

Instead of manual text prompt inputs $P_m$, we leverage an image sequence captioning model to generate text prompts for the input image sequence, addressing the divergence in image interpretation between humans and captioning models. Fig.~\ref{fig:lpe} shows that while coarse-grained text prompts $P_c$ from previous text-to-image generation models can yield correct person images, they lack ID consistency. Manual text prompts $P_m$ often carry individual subjective nuances, occasionally lacking precision, overlooking certain features, or being overly specific, resulting in a lack of variety in the generated images. Since the textual output from the multimodal model aligns most consistently with the model's understanding of the image, we decided to use the text prompts $P_g$ generated by caption models. After extensive experiments (shown in Sec.\ref{5.3}), we select BLIP2~\cite{li2023blip} as the captioning model since it exhibits robust image-to-text generation capability. To make it suitable for person image generation, we adjust the attention of BLIP2 to focus on the attributes of person images, such as carried items and actions.

Text-to-image diffusion models typically leverage pre-trained vision-language models (e.g., CLIP~\cite{radford2021learning}) to obtain well-matched text embeddings and image embeddings in the latent space. Following previous methods, we adopt a pre-trained text encoder from CLIP to embed input text prompts. During fine-tuning, we introduce an an Identity Information Representer (IIR) as part of the text prompt, which is a meaningless text not in CLIP's vocabulary. IIR enforces a strong relationship between itself and the identity information of the input image sequence while preserving the pre-trained mapping between text embedding and image embedding. Notably, fine-tuning the text-to-image diffusion model without IIR results in the model learning coarse-grained category information rather than fine-grained identity details. 
On the contrary, incorporating IIR during fine-tuning tends to lead the model towards overfitting, where it excessively matches the image embedding of the input image sequence with the text embedding of IIR in the feature space. This overfitting is essential for guaranteeing ID consistency between the generated and input images.

\textbf{Diversity Injection(DI)}. As mentioned in Section~\ref{sec1}, we rethink the two requirements for generating data in person Re-ID task, ID consistency and attribute diversity. Attribute diversity ensures the backbone encounters numerous appearances during pre-training, enhancing its generalization capabilities in the subsequent fine-tuning phase. The DI module aims to introduce novel attributes not present in the input image sequence, such as new clothing, carried items, actions, camera poses, and scenes.

We utilize consecutive image sequences from the same camera as inputs and generate a \textit{Attribute Reference Set} (200 images in total), using a pre-trained Stable Diffusion~\cite{rombach2022high} that adopts LPE prompts without IIR . The generation process here is coarse-grained, aiming to broadly generate pedestrian images to obtain a greater diversity of features. To fine-tune the diffusion model, we propose the Fine-Grain-Specific Prior Preservation Loss based on \textit{Attribute Reference Set}, which can be formulated as:

\begin{equation}
\begin{aligned} 
\mathbb{E}_{\mathbf{x}, \mathbf{c}, \boldsymbol{\epsilon}, \boldsymbol{\epsilon}^{\prime}, t}\left[w_{t}\left\|\hat{\mathbf{x}}_{\theta}\left(\alpha_{t} \mathbf{x}+\sigma_{t} \boldsymbol{\epsilon}, \mathbf{c}\right)-\mathbf{x}\right\|_{2}^{2}+\right.  \left.\lambda w_{t^{\prime}}\left\|\hat{\mathbf{x}}_{\theta}\left(\alpha_{t^{\prime}} \mathbf{x}_{\mathrm{crs}}+\sigma_{t^{\prime}} \boldsymbol{\epsilon}^{\prime}, \mathbf{c}_{\mathrm{LPE*}}\right)-\mathbf{x}_{\mathrm{crs}}\right\|_{2}^{2}\right]
\end{aligned}
\label{eq5}
\end{equation}

Eq.~\ref{eq5} contains two items. The first item is to supervise the text-to-image processing. $\mathbf{X}$ represents the input image; $\epsilon$ denotes the Gaussian noise; $c$ is the text condition; $\hat{\mathbf{x}}_\theta$ represents the diffusion model. Following previous works, we also adopt L2 loss to supervise generation. The second item is the prior-preservation term that supervises the model with its own generated images: $x_{\mathrm{crs}}=\hat{x}\left(\mathbf{z}, \mathbf{c}_{\mathrm{LPE*}}\right)$ is the Attribute Reference Set (200 images in total); $\mathbf{c}_{\mathrm{LPE*}}$ represents the enhanced text condition through LPE. $\lambda$ balances the weights of two terms. 

\subsection{Filtering Stage}
\label{4.2}
In the generation stage, numerous person images can be obtained. However, it is inevitable to encounter low-quality samples due to the large feature space of the diffusion model. To tackle this problem, we experimented with three candidate filtering approaches to filter out those low-quality samples:

$\bullet$\textbf{CLIP Filter}. We employ a classification model by CLIP to filter generated images, selecting images equipped with high confidence scores and discarding those with low confidence scores.

$\bullet$\textbf{Classification Confidence Threshold Filter (CCTF)}. We train an image classification model based on person identity using source images and remove generated images with low classification confidence. 

$\bullet$\textbf{Re-ID Confidence Threshold Filter (Re-ID CTF)}. We independently train person Re-ID models separately for each of the four source datasets (Market-1501~\cite{zheng2015scalable}, Airport~\cite{karanam2016comprehensive}, MSMT17~\cite{wei2018person} and CUHK03~\cite{li2014deepreid}) used in the generation process. We evaluate the generated images based on their confidence scores to ascertain successful matching with the source images. Images with higher confidence scores are selected, and lower ones are discarded.

The CLIP filter is a valuable tool for assessing text-to-image generation models' accuracy. It delivers excellent filtering results in coarse-grained generation tasks. However, in our fine-grained generation task that emphasizes ID consistency, the CLIP filter's confidence score primarily reflects the accuracy of category generation rather than the consistency of identities. CCTF effectively selects generated images that exhibit high visual similarity to the source image. It enables the removal of low-quality generated images, such as those exhibiting significant style variations or missing generation subjects. However, CCTF focuses more on the images' overall visual features than the person's specific features. For example, we appreciate a diversity of scenes in our images, but CCTF may assign low confidence scores to these images due to their varied vision feature. Re-ID CTF compares the Re-ID features of generated images to those of the source images, filtering the generated images based on their similarity confidence scores. It is more aligned with our downstream tasks and is better equipped to select images beneficial for pre-training. Fig.~\ref{fig:filter} shows some filtered results based on these three candidate filtering approaches. CLIP Filter keeps images containing ``person''. However, they do not have ID consistency between the input image sequence and generated images, as they deliver filtering results in coarse-grained generation tasks. CCTF reserves only images that are highly similar in overall visual features rather than the person's specific features. Re-ID CTF can be selected with ID consistency and attribute diversity that meet our requirements.

\begin{table*}[t]
    \centering
    \small
    \scalebox{0.85}{
    \begin{tabular}{c c c c c c c}
        \toprule
        Datasets & images & scene & person IDs & labeled & environment  & crop size \ \\
        \midrule
        VIPeR~\cite{gray2008viewpoint} & 1,264 & 2 & 632 & yes & -  & $128\times48$ \\
        GRID~\cite{liu2012person} & 1,275 & 8 & 1,025 & yes & subway  & vary \\
        CUHK03~\cite{li2014deepreid} & $14,096$ & $2$ & $1,467$ & yes & campus  & vary\\
        Market1501~\cite{zheng2015scalable} & $32,668$ & $6$ & $1,501$ & yes & campus  & $128\times64$ \\
        Airport~\cite{karanam2016comprehensive} & $39,902$ & $6$ & $9,651$ & yes & airport  & $128\times64$ \\
        % DeepChange~\cite{xu2021long} & $\simeq187,000$ & $17$ & $1,121$ & yes & open street view  &  vary \\
        MSMT17~\cite{wei2018person} & $126,441$ & $15$ & $4,101$ & yes & campus  & vary \\
        \midrule
        \ ImageNet-1K~\cite{imagenet15russakovsky} & $1,281,167$ &  -  & $1k$(not person)  & yes & -  & vary \\
         LUPerson~\cite{fu2021unsupervised} & $4,180,243$ & $46,260$ & $>200k$ & no & vary  & vary \\
         SYSU30K~\cite{wang2020weakly} & 29,606,918 & 1,000 & 30,508 & weakly & TV program  & vary \\
         LUPerson-NL~\cite{fu2022large} & $10,683,716$ & $21,697$ & $\simeq433,997$ & noisy & vary  & vary \\
        \midrule
        PersonX~\cite{sun2019dissecting} & $273,456$ & $12$ & $1,260$ & yes & virtual environments  & vary \\
        RandPerson~\cite{wang2020surpassing} & $132,145$ & $11$ & $8,000$ & yes & virtual environment  & vary \\
        \midrule
        \textbf{Diff-Person} & $777,130$ & vary & 5,183 & yes & vary  & $256\times128$ \\
        \bottomrule
    \end{tabular}}

        \caption{From top to bottom are supervised, pre-training, and synthetic (based on virtual environments) datasets for person Re-ID.}
    \label{tab:data-stat}

\end{table*} 

\begin{wrapfigure}{R}{0.45\columnwidth}
    \vspace{-0.15in}
    \centering
    \includegraphics[width=0.45\textwidth]{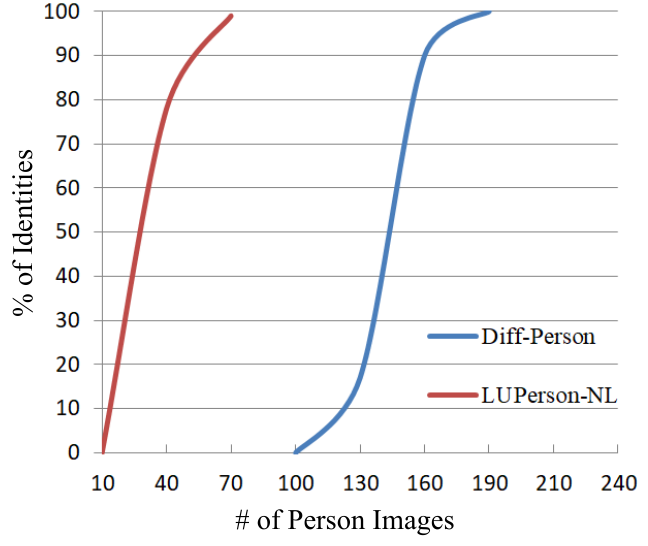}
    \vspace{-0.2in}
\caption{\textbf{Identity distribution of \textit{Diff-Person} and LUPerson-NL.} A curve point(X, Y) indicates Y\% of identities each has less than X images.}
    \label{fig:distribution}
    \vskip -0.15in
\end{wrapfigure}

\subsection{Diff-Person: A Novel Re-ID Dataset for Pre-training }
\label{4.3}
In this work, we utilize \textit{Diffusion-ReID} to establish a large number of annotated person images using only the training set of four widely-used supervised person Re-ID datasets (Market-1501~\cite{zheng2015scalable},  MSMT17~\cite{wei2018person}, CUHK03~\cite{li2014deepreid} and Airport~\cite{karanam2016comprehensive}). To our knowledge, this is the first person Re-ID dataset generated by diffusion models. \textit{Diff-Person} currently consists of more than 777K images from 5,183 identities.

We detail the statistics of existing popular person Re-ID datasets in Tab.~\ref{tab:data-stat}. Compared to existing annotated datasets, \textit{Diff-Person} contains a large number of images, scenes, and person IDs, significantly surpassing supervised datasets. Compared to MSMT17, a widely-used large-scale dataset, \textit{Diff-Person} holds more than six times the number of images from more person identities. Regarding scene diversity, \textit{Diff-Person} has a clear advantage due to its personalized attributes generation, which can generate scenes and camera poses that do not appear in the input image sequence. As for existing synthetic datasets PersonX and RandPerson, generated by manual modeling, they also introduce a domain gap between the virtual environment and the real world. Training models based on these datasets cannot be effortlessly migrated to real applications.

\begin{figure*}[t]
    \centering
    \includegraphics[width=0.8\linewidth]{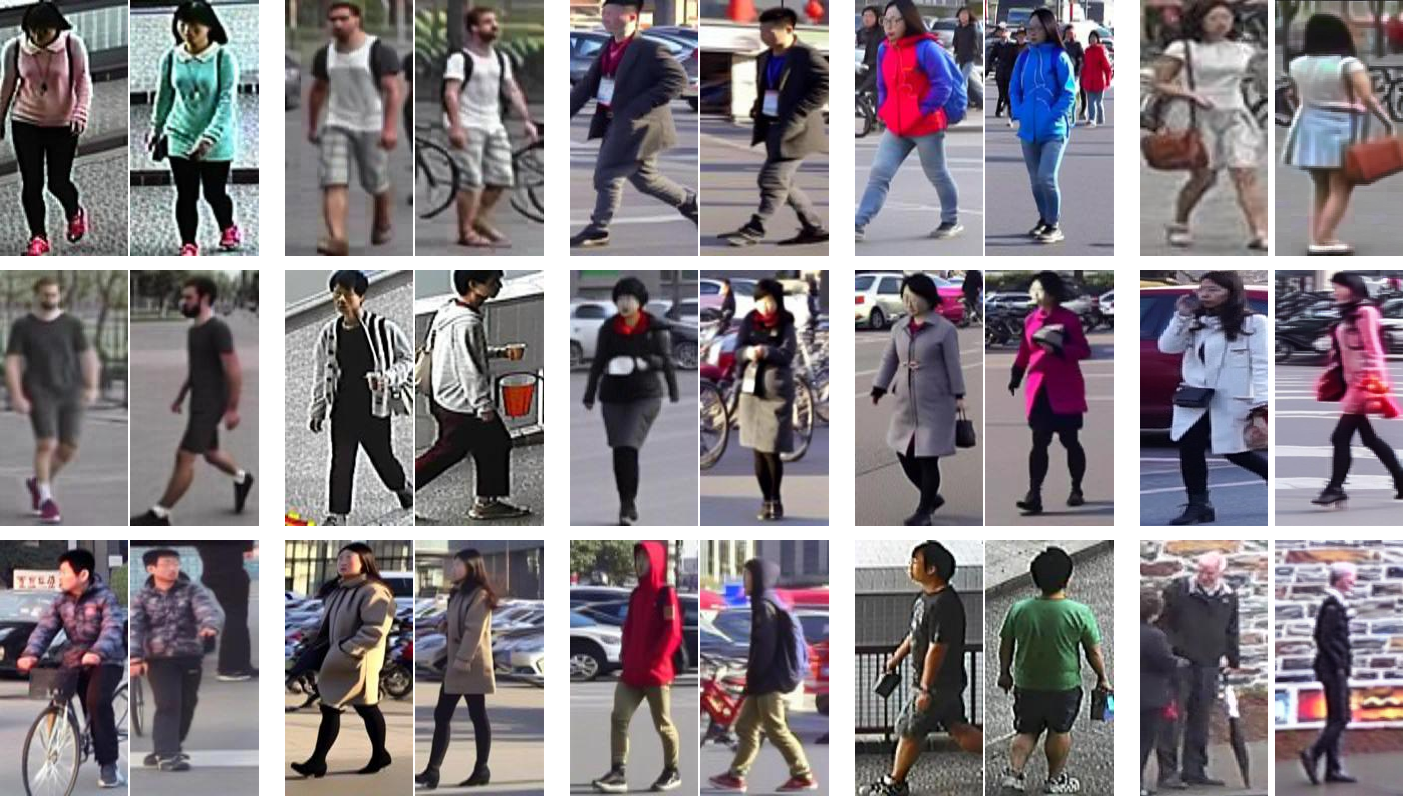}
\caption{
\textbf{More example images generated by \textit{Diff-Person}.}
}
\label{fig:sample}
\end{figure*}

ImageNet-1K is currently the most widely used pre-training dataset. Even though the pre-trained backbone on ImageNet-1K has shown promising performance in person Re-ID, it is still necessary to address the issue of establishing a better dataset for pre-training in person Re-ID due to the significant domain gap between ImageNet-1K and person Re-ID datasets. However, large-scale unsupervised or weakly supervised person Re-ID datasets such as LUPerson and SYSU30K have some shortcomings: 1) They do not have explicit ID labels or treat trajectory labels as noisy ID labels, which is not highly accurate to person Re-ID tasks; 2) They require significant time costs and human resources, even for unsupervised datasets, which require much time spent collecting data and filtering out invalid samples; 3) Since the data of these datasets come from various online channels, their distribution usually has the problem of long-tail distribution, which is not conducive to pre-training. Compared with existing person Re-ID datasets, \textit{Diff-Person} has explicit annotated data with diversity attributes that can be generated efficiently and cost-effectively using existing annotated person Re-ID datasets. Moreover, the data distribution is uniform, which is more beneficial to obtaining a better pre-training backbone. We illustrate the cumulative percentage of identities concerning the number of their corresponding person images as a curve in Fig.~\ref{fig:distribution}. A point (X, Y) on the curve represents the total Y\% identities in \textit{Diff-Person} and LUPerson-NL~\cite{fu2022large}, and each has less than X images. It can be observed that: 1) Approximately 80\% of the identities in \textit{Diff-Person} are associated with a number of person images falling within the range of [70, 210], which is much larger than LUPerson-NL; 2) The percentage of identities that have more than 130 person images occupy about 70\% in \textit{Diff-Person}. These observations show that our \textit{Diff-Person} is more beneficial to obtaining a better pre-training backbone. Fig.~\ref{fig:sample} shows more example images in \textit{Diff-Person}.

\section{Experiments}\label{sec5}

\subsection{Implementation Details}
When fine-tuning the stable diffusion model v1.4 for each specific ID, we resize the input images and condition maps to 512 × 512. We set the learning rate to 2e-6, and the training step is 1000. The training process for 5k IDs is performed on 8 NVIDIA GeForce RTX 4090 GPUs and can be completed within three days. In inference, the proposed method can generate ten images within 15 seconds with a single NVIDIA GeForce RTX 4090 in parallel.
 
During pre-training, we train ResNet-50 for 300 epochs. The AdamW optimizer is adopted with a learning rate of 4e-3 and a weight decay of 0.05. There is a 20-epoch linear warm-up and a cosine decaying schedule afterward. We use a batch size of 512. For data augmentation, we adopt common schemes including Mixup, Cutmix, RandAugment, and Random Erasing.

In our work, we fine-tune the pre-trained backbone on four widely used datasets (Market-1501~\cite{zheng2015scalable},  MSMT17~\cite{wei2018person}, CUHK03~\cite{li2014deepreid} and Airport~\cite{karanam2016comprehensive}) to evaluate its advantages over the pre-trained backbone by ImageNet-1K.

\subsection{Experimental Results}
\label{sec:5.2}
We compare the performance of our pre-trained backbone and ImageNet-1K pre-trained backbone in six settings: the supervised person Re-ID setting, the few-shot Re-ID settings, the unsupervised Re-ID setting, the domain adaption Re-ID setting, and the domain generalization Re-ID setting. We utilize the commonly used metrics in person Re-ID, such as Cumulative Match Characteristic (CMC) and Mean Average Precision (mAP). All experimental results are derived from re-implementations using corresponding open-source code. To ensure a fair comparison, identical parameters are applied in the training process for both backbones. All results in Section~\ref{sec5} are shown in mAP/Rank-1.

\textbf{Evaluation for Supervised Learning}.
The supervised person Re-ID setting is one of the most widely used experimental setups.
To compare the performance of the \textit{Diff-Person} pre-trained backbone and the ImageNet-1K pre-trained backbone, we use four representative methods: MGN~\cite{Wang_Yuan_Chen_Li_Zhou_2018}, AGW~\cite{pami21reidsurvey}, SBS~\cite{he2020fastreid}, and BDB~\cite{dai2019batch}. Our pre-trained backbone improves performance for all four baseline methods on four
person Re-ID benchmarks. Specifically, Tab.~\ref{tab:impro-sup} shows that the most significant performance improvement is for MGN on CUHK, where mAP is improved by 9.4\%, and Rank-1 improves by 10.1\%.
More notably, AGW increases mAP of 2.6\%, 2.9\%, 3.3\%, and 5.5\% on CUHK03, Market-1501, PersonX, and MSMT17, respectively. Regarding Rank-1, AGW also achieves improvements of 3.4\%, 1.2\%, 0.8\%, and 5.0\%. Despite the impressive results already achieved on Market-1501 by existing methods, our pre-trained backbone delivers a noteworthy improvement. Moreover, our experiments showcase enhanced performance on the synthetic dataset PersonX and the larger-scale MSMT17 dataset, emphasizing the significant advantages of our pre-trained backbone over those pre-trained on ImageNet-1K.
However, the performance gain on MSMT17 is less pronounced than on other datasets. A plausible explanation is that the substantial data in MSMT17 makes it sufficient for fine-tuning a more robust backbone.

\begin{table*}

    \centering
            \caption{Comparing four supervised Re-ID methods using ImageNet-1K pre-trained backbone and \textit{Diff-Person} pre-trained backbone. 
    ``IN'' and ``Ours'' indicate that the backbone is pre-trained on ImageNet-1K and \textit{Diff-Person}, respectively.}
    \scalebox{0.8}{
   
    \begin{tabular}{cccccc}
    \toprule    
    Dataset & Method &MGN~\cite{Wang_Yuan_Chen_Li_Zhou_2018}  & AGW~\cite{pami21reidsurvey} & SBS~\cite{he2020fastreid} & BDB~\cite{dai2019batch}\\
 
    \midrule
    \multirow{3}{*}{CUHK03~\cite{li2014deepreid}} & IN & 66.0/66.8 & 62.0/63.6 & 65.0/68.9 & 69.3/72.8  \\
     & \multirow{2}{*}{Ours} & \textbf{75.4/76.9} &\textbf{64.6/67.0} &\textbf{69.4/70.0} &\textbf{75.3/77.3} \\
      &   & \textcolor{teal}{+9.4/+10.1} &\textcolor{teal}{+2.6/+3.4} &\textcolor{teal}{+4.4/+1.1} &\textcolor{teal}{+6.0/+4.5} \\
    \midrule
    \multirow{3}{*}{Market-1501~\cite{zheng2015scalable}} & IN &86.9/95.7 &87.8/95.1 &88.2/95.4 &84.3/94.2 \\
     & \multirow{2}{*}{Ours} & \textbf{88.2/96.3} &\textbf{90.7/96.3}&\textbf{89.4/95.8} &\textbf{85.2/94.7} \\
     & & \textcolor{teal}{+1.3/+0.6} & \textcolor{teal}{+2.9/+1.2} & \textcolor{teal}{+1.2/+0.4}&\textcolor{teal}{+0.9/+0.5}\\
     \midrule
    \multirow{3}{*}{PersonX~\cite{sun2019dissecting}} & IN &85.3/94.3 &87.9/95.8 &90.1/95.3 &84.4/89.8 \\
     & \multirow{2}{*}{Ours} & \textbf{87.6/94.6}&\textbf{89.2/96.6}&\textbf{92.3/95.5} &\textbf{87.4/91.1} \\
     & & \textcolor{teal}{+2.3/+0.3} & \textcolor{teal}{+3.3/+0.8} & \textcolor{teal}{+2.2/+0.2}&\textcolor{teal}{+3.0/+1.3}\\
     \midrule
    \multirow{3}{*}{MSMT17~\cite{wei2018person}} & IN &65.4/85.1 &49.3/68.3 &58.4/81.8 &69.3/72.8 \\
     & \multirow{2}{*}{Ours} & \textbf{66.1/85.4} &\textbf{54.8/73.3} &\textbf{60.5/81.9} &\textbf{69.5/73.4}\\
     & & \textcolor{teal}{+0.7/+0.3} & \textcolor{teal}{+5.5/+5.0} & \textcolor{teal}{+2.1/+0.1}&\textcolor{teal}{+0.2/+0.6}\\
    \bottomrule
    \end{tabular}}
    
    \label{tab:impro-sup}
\end{table*}

\begin{wrapfigure}{R}{0.6\columnwidth}
    \vspace{-0.15in}
    \centering
    \includegraphics[width=0.6\textwidth]{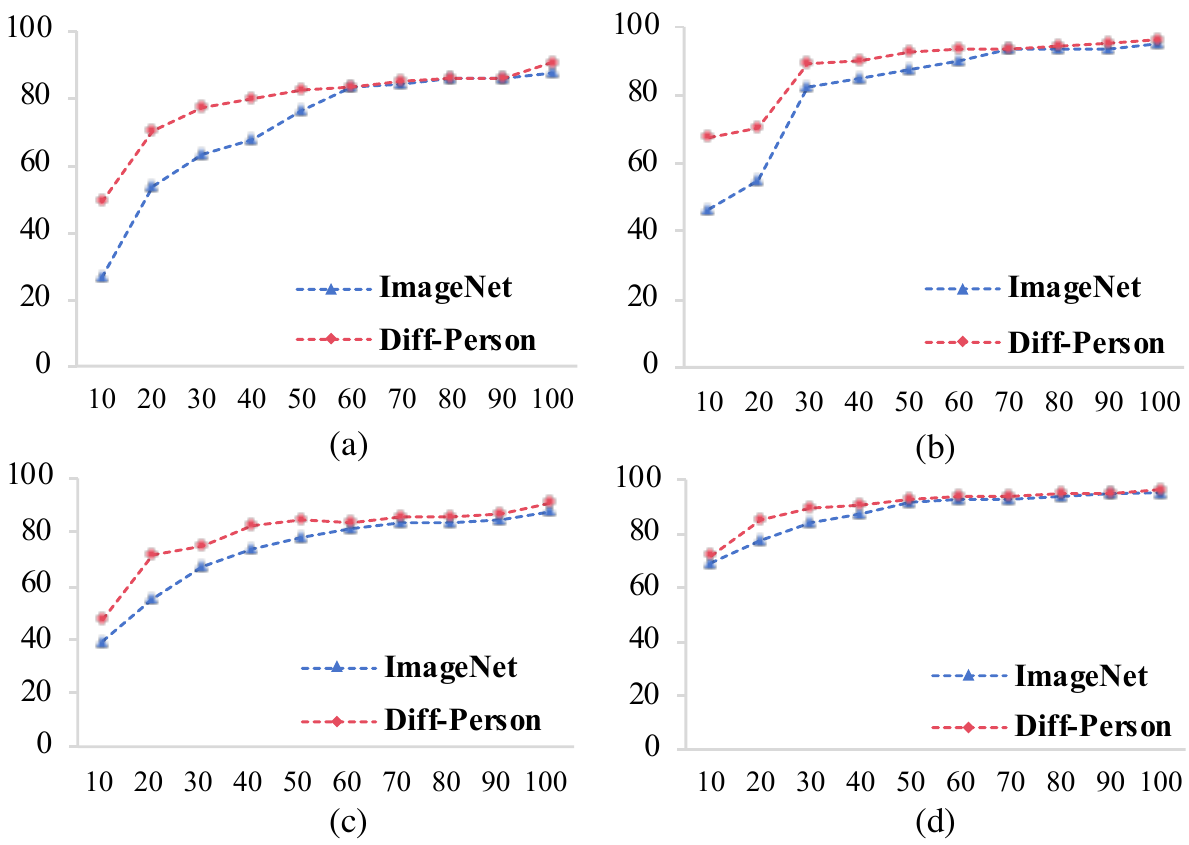}
    \vspace{-0.2in}
    \caption{\textbf{Performance comparison in the few-shot image setting and the small-scale identity setting.} (a) and (b) represent the trends as the data scale incrementally increases from 10\% of the full dataset for the few-shot image setting; (c) and (d) represent the trends as the data scale incrementally increases from 10\% of the full dataset for the small-scale identity setting. }
    \label{fig:fssc}
    \vskip -0.15in
\end{wrapfigure}

\textbf{Evaluation for Few-shot Learning}.
\label{fs}
We adhere to the scheme proposed in LUPerson~\cite{fu2021unsupervised}, 
conducting experiments in two few-shot learning settings, the few-shot image setting, and the small-scale identity setting. Concretely, the few-shot image setting limits the available percentage of images for each identity, whereas the small-scale identity setting confines the percentage of accessible identities.
These constraints hold paramount significance in real-world person Re-ID applications. In practical scenarios, acquiring sufficient data that satisfies a substantial number of images and a diverse range of pedestrian identities poses a challenge. Consequently, these settings confirm that our pre-trained backbone is superior in extracting robust features for practical applications. Within these settings, we adjust the available data percentage of Market-1501, ranging from 10\% to 90\%. We use AGW as our baseline method.
Figure~\ref{fig:fssc} illustrates the performance trend of AGW when using varying percentages of data under the few-shot image setting and the small-scale identity setting, employing backbones pre-trained on both \textit{Diff-Person} and ImageNet-1K. Notably, our pre-trained backbone consistently outperforms the ImageNet-1K pre-trained backbone, with the advantages becoming more pronounced as the labeled data becomes more limited. Detailed results in Table~\ref{tab:fs} confirm these findings. In the few-shot image setting, our backbone surpasses the ImageNet-1K pre-trained backbone by 22.5\% in mAP and 21.0\% in Rank-1 accuracy, with only 10\% of the images available for each identity. Furthermore, our pre-trained backbone nearly matches the performance of the ImageNet-1K pre-trained backbone at 90\% data, utilizing only 50\% of the available data. This outstanding performance in the few-shot learning scenarios underscores the practical value of our pre-trained backbone for real-world person Re-ID applications.

\begin{table*}[t]
    \centering
    \caption{Performance comparison with different percentages of training data used for fine-tuning for the few-shot image and the small-scale identity settings. ``Fs'' indicates the few-shot image setting, and ``Ss'' indicates the small-scale identity setting.}
    \scalebox{0.8}{
    \begin{tabular}{ccccccccccc}
    \toprule    
    Setting & Method & $10\%$ & $30\%$& $50\%$ & $70\%$ & $90\%$\\
    \midrule
    \multirow{3}{*}{Fs} & IN & 26.4/46.5 & 63.0/82.4  & 76.7/87.5  & 84.2/93.4  & 85.5/93.8 \\
      & \multirow{2}{*}{Ours} & \textbf{48.9/67.5} & \textbf{77.4/89.6} & \textbf{82.7/92.9} & \textbf{84.9/93.8} & \textbf{86.3/94.8}\\
      & &\textcolor{teal}{+22.5/+21.0}  &\textcolor{teal}{+14.4/+7.2} &\textcolor{teal}{+6.0/+5.4}&\textcolor{teal}{+0.7/+0.4}  &\textcolor{teal}{+0.8/+1.0}\\
    \midrule
    \multirow{3}{*}{Ss} & IN & 39.3/69.1 & 67.4/84.0 & 78.0/92.0 & 83.6/93.2 & 85.0/94.3\\
      & \multirow{2}{*}{Ours} & \textbf{47.3/71.6}  & \textbf{75.0/89.8} & \textbf{84.5/93.0} & \textbf{85.7/93.7} & \textbf{86.7/94.5}\\
            & &\textcolor{teal}{+8.0/+2.5} &\textcolor{teal}{+7.6/+5.8} &\textcolor{teal}{+6.5/+1.0} &\textcolor{teal}{+2.1/+0.5} &\textcolor{teal}{+1.7/+0.2}\\
    \bottomrule
    \end{tabular}}
    \label{tab:fs}
\end{table*}

\begin{table*}
    \centering
            \caption{Comparing unsupervised person Re-ID setting. }
    \scalebox{0.8}{
    
    \begin{tabular}{cccccccccccc}
    \toprule    
 Dataset & Method &C-Contrast~\cite{dai2022cluster} & PPLR~\cite{cho2022part} &HHCL~\cite{hu2021hard}&SpCL~\cite{ge2020self}&ICE~\cite{chen2021ice} \\
    \midrule
 \multirow{3}{*}{Market-1501~\cite{zheng2015scalable}} & IN    &82.6/93.0  &84.4/94.3 &84.2/93.4 &72.6/87.7 &82.3/93.8\\ 
    & \multirow{2}{*}{Ours} & \textbf{85.3/93.8} & \textbf{86.7/95.2} & \textbf{85.9/94.9}& \textbf{80.2/91.7} & \textbf{84.3/94.9}\\
     & & \textcolor{teal}{+2.7/+0.8} &\textcolor{teal}{+2.3/+0.9} &\textcolor{teal}{+1.7/+1.5} &\textcolor{teal}{+7.6/+4.0}&\textcolor{teal}{+2.0/+1.1}\\
     \midrule
         \multirow{3}{*}{MSMT17~\cite{wei2018person}} & IN  & 33.3/63.3 & 42.2/73.3 & 40.1/69.8 & 25.3/57.8 &38.9/70.2\\
     & \multirow{2}{*}{Ours} & \textbf{40.4/68.7}  & \textbf{45.6/75.5}& \textbf{42.5/73.9}& \textbf{32.5/63.4}& \textbf{43.5/73.1}\\
     & & \textcolor{teal}{+7.1/+5.4} & \textcolor{teal}{+3.3/+2.2} & \textcolor{teal}{+2.4/+4.1}& \textcolor{teal}{+7.2/+5.6}& \textcolor{teal}{+4.6/+2.9}\\
    \bottomrule
    \end{tabular}}
    \label{tab:impro-unsup}
\end{table*}

\textbf{Evaluation for Unsupervised Learning}.
Unsupervised person Re-ID aims to train a deep model with the capacity to identify individuals of interest within extensive unlabeled datasets. 
We compare our pre-trained backbone with the ImageNet-1K pre-trained backbone in the unsupervised person Re-ID setting. We select C-Contrast~\cite{dai2022cluster}, PPLR~\cite{cho2022part}, HHCL~\cite{hu2021hard}, SpCL~\cite{ge2020self}, and ICE~\cite{chen2021ice} as the baseline models. The results presented in Tab.~\ref{tab:impro-unsup} reveal that our pre-trained backbone consistently outperforms the ImageNet-1K pre-trained backbone. Notably, C-Contrast achieves improvements of 2.7\% and 7.1\% mAP on Market-1501 and MSMT17, respectively. In the case of MSMT17, known for its larger data volume and increased difficulty, all methods attain superior mAP and Rank-1 results. It underscores the superior robust feature extraction capabilities of our pre-training backbone.

\textbf{Evaluation for Domain Adaption}.
Domain adaption person Re-ID aims to transfer acquired knowledge from a labeled source domain to an unlabeled target domain. The limited generalization of person Re-ID models trained on specific domains due to the diversity of data access scenarios hampers their applicability in real-world situations. 
We compare our pre-trained backbone in the domain adaptation person Re-ID setting with the ImageNet-1K pre-trained backbone. Five baseline models are selected for experiments, including CaCL~\cite{lee2023camera}, IDM~\cite{dai2021idm}, HCD~\cite{zheng2021online}, GLT~\cite{zheng2021group}, and UNRN~\cite{zheng2021exploiting}. The results in Tab.~\ref{tab:da} demonstrate that our method outperforms the baseline model, achieving improvements of 1.7\%/0.9\% and 1.9\%/0.3\% in comparison to the SOTA method CaCL on Market-1501 to MSMT17 (MA2MS) and MSMT17 to Market-1501 (MS2MA), respectively. These results underscore the effectiveness of our pre-trained backbone in delivering superior initialization for domain adaption in person Re-ID.
\begin{table*}
    \centering
            \caption{Comparing domain adaption person Re-ID setting.}
    \scalebox{0.8}{

    \begin{tabular}{cccccccccccc}
    \toprule    
 Dataset & Method &CaCL~\cite{lee2023camera}  & IDM~\cite{dai2021idm}&HCD~\cite{zheng2021online}&GLT~\cite{zheng2021group} &UNRN~\cite{zheng2021exploiting}\\

    \midrule
 \multirow{3}{*}{MA2MS} & IN    &36.5/66.6  &33.5/61.3 &28.4/54.9 &26.5/56.6 &25.3/52.4\\ 
    & \multirow{2}{*}{Ours} & \textbf{38.2/67.5} & \textbf{34.7/61.6} & \textbf{32.4/59.7}& \textbf{31.7/60.4}& \textbf{31.4/58.9}\\
     & & \textcolor{teal}{+1.7/+0.9} &\textcolor{teal}{+1.2/+0.3} &\textcolor{teal}{+4.0/+4.8} &\textcolor{teal}{+5.2/+3.8} &\textcolor{teal}{+6.1/+6.5}\\
     \midrule
         \multirow{3}{*}{MS2MA} & IN  & 84.7/93.8 & 82.1/92.4 & 80.2/91.4 & 79.3/90.7& 78.3/90.4\\
     & \multirow{2}{*}{Ours} & \textbf{86.6/94.1}  & \textbf{84.8/93.9}& \textbf{82.3/92.5}& \textbf{81.7/92.2}& \textbf{81.4/92.0}\\
     & & \textcolor{teal}{+1.9/+0.3} & \textcolor{teal}{+2.7/+1.5} & \textcolor{teal}{+2.1/+1.1}& \textcolor{teal}{+2.4/+1.5}& \textcolor{teal}{+3.1/+1.6}\\
    \bottomrule
    \end{tabular}}
    \label{tab:da}
\end{table*}

\begin{table*}
    \centering
    \caption{Comparing domain generalization person Re-ID setting.}
    \scalebox{0.8}{
    \begin{tabular}{cccccccccccc}
    \toprule    
 Method & Backbone &PRID~\cite{hirzer2011person}  & GRID~\cite{liu2012person}&VIPeR~\cite{gray2008viewpoint}&iLIDS~\cite{Zheng_Gong_Xiang_2009} &Average\\
    \midrule
 \multirow{3}{*}{MetaBIN~\cite{choi2021meta}} & IN    & 81.0/74.2 & 57.9/48.4 & 68.6/59.9 & 87.0/81.3 & 73.6/66.0 \\ 
    & \multirow{2}{*}{Ours} &\textbf{85.7/79.3} & \textbf{61.2/52.0} & \textbf{72.1/63.6} & \textbf{90.2/85.0} & \textbf{77.0/72.9}\\
     & & \textcolor{teal}{+4.7/+5.1}  & \textcolor{teal}{+3.3/+3.7}&\textcolor{teal}{+3.5/+3.7}& \textcolor{teal}{+3.2/+3.7}& \textcolor{teal}{+3.4/+5.9}\\
     \midrule
         \multirow{3}{*}{META~\cite{xu2022mimic}} & IN  & 71.7/61.9 & 57.9/50.2 & 64.3/55.9 & 81.5/74.0 & 66.8/57.3\\
     & \multirow{2}{*}{Ours} &\textbf{79.3/77.0} & \textbf{61.5/52.0} & \textbf{71.9/63.9} & \textbf{83.5/79.2} & \textbf{74.5/68.0}\\
     & & \textcolor{teal}{+7.6/+5.1}  & \textcolor{teal}{+3.6/+1.8}&\textcolor{teal}{+7.6/+8.0}& \textcolor{teal}{+2.0/+5.2}& \textcolor{teal}{+7.7/+10.7}\\
    \bottomrule
    \end{tabular}}
    \label{tab:gene}
\end{table*}

\textbf{Evaluation for Domain Generalization}.
Domain generalization person Re-ID endeavors to cultivate a robust model capable of delivering optimal performance on an unseen target domain without additional updates. Domain generalization person Re-ID proves more practical and challenging by leveraging training data from multiple source domains and conducting direct testing across diverse and unseen domains. This distinction has garnered increasing attention due to its real-world applicability.
Following the experimental setting outlined in META~\cite{xu2022mimic}, we utilize all images from the source domains (Market-1501, CUHK02~\cite{li2013locally}, CUHK03, and CUHK-SYSU~\cite{xiao2016end}) for training. For target sets (PRID~\cite{hirzer2011person}, GRID~\cite{liu2012person}, VIPeR~\cite{gray2008viewpoint}, and iLIDS~\cite{Zheng_Gong_Xiang_2009}), the results are evaluated based on the average of 10 repeated random splits of query and gallery sets.
We employ MetaBIN~\cite{choi2021meta} and META~\cite{xu2022mimic} as the baseline methods. As depicted in Tab.~\ref{tab:gene}, our backbone consistently outperforms the ImageNet-1K pre-trained backbone. Specifically, META with the \textit{Diff-Person} pre-trained backbone exhibits an average increase of 7.7\% in mAP and 10.7\% in Rank-1. These results underscore the effectiveness of our pre-trained backbone in providing superior initialization for domain generalization in person Re-ID.

\subsection{Ablation Study}
\label{5.3}
In the LPE module, we utilize an image sequence captioning model to generate text prompts of the input image sequence, mitigating the discrepancy in image interpretation between humans and captioning models. Our experiments involve three image caption models—VLP~\cite{li2022blip}, BLIP~\cite{li2022blip}, and BLIP2~\cite{li2023blip}. We apply our paradigm to conduct generation experiments based on these three image sequence captioning models. The experimental setting is the same as the supervised person Re-ID.
As shown in Tab.~\ref{tab:captionmodel}, the results show that BLIP2 achieved the best performance over other methods. We attribute the performance improvement to BLIP2's enhanced alignment of visual and textual feature spaces, making it a more suitable choice for our specific requirements. Therefore, we chose BLIP2 as the image sequence captioning model in the LPE module.
\begin{table*}
    \centering
    \caption{Comparison of different image sequence captioning models.}
    \scalebox{0.8}{
    \setlength{\tabcolsep}{2.8mm}{
    \begin{tabular}{ccccc}
    \toprule    
    Market-1501 & MGN~\cite{Wang_Yuan_Chen_Li_Zhou_2018}  & AGW~\cite{pami21reidsurvey}  & SBS~\cite{he2020fastreid}  & BDB~\cite{dai2019batch}  \\ \midrule
        VLP~\cite{li2022blip} & 83.1/87.6 & 84.2/89.7 & 83.9/88.4 & 77.5/79.9  \\ \midrule
        BLIP~\cite{li2022blip} & 85.1/92.1 & 87.2/92.1 & 86.2/91.4 & 80.7/89.2  \\ \midrule
        BLIP2~\cite{li2023blip} & 88.2/96.3 & 90.7/96.3 & 89.4/95.8 & 85.2/94.7  \\ 
    \bottomrule
    \end{tabular}}}
    \label{tab:captionmodel}
\end{table*}
\subsection{Discussions}
\label{5.4}

\begin{table*}
    \centering
    \caption{Comparison of different pre-training data scales. ``Image Scale'' indicates the different scales of image quantity, ``ID Scale'' indicates the different scales of person ID quantity.}
    \scalebox{0.74}{
    \begin{tabular}{ccccccccc}
    \toprule    
    \multirow{2}{*}{Dataset} & \multicolumn{4}{c}{Image Scale}& \multicolumn{4}{c}{ID Scale}  \\
    \cmidrule{2-9}
     &   $12.5\%$ & $25\%$ & $50\%$ & $100\%$  &   $12.5\%$ & $25\%$ & $50\%$ & $100\%$\\
    \midrule
    CUHK03~\cite{li2014deepreid} &64.3/65.7 &65.0/67.5 &69.2/73.9 &75.4/76.9  &69.9/69.3 &73.9/74.7 &75.2/76.2 &75.4/76.9 \\
     \midrule
    Market-1501~\cite{zheng2015scalable} &84.9/94.2 &86.1/94.7 &87.6/95.2 &88.2/96.3 &86.9/94.4 &87.2/95.2 &87.9/96.0 &88.2/96.3  \\
     \midrule
    PersonX~\cite{sun2019dissecting} & 85.7/93.2 &86.2/93.9 &87.2/94.2 &87.6/94.6 & 85.4/94.1 &86.2/94.3 &87.5/94.6 &87.6/94.6 \\
     \midrule
    MSMT17~\cite{wei2018person} & 59.3/80.5 &62.2/82.8 &64.9/84.1 &66.1/85.4 & 61.2/82.7 &63.7/83.4 &65.4/85.0 &66.1/85.4 \\
    \bottomrule
    \end{tabular}}
    \label{tab:imagescale}
\end{table*}

\textbf{The Effectiveness of Pre-training Data Scale}. 
As we analyzed in Evaluation for the few-shot Learning, the number of person IDs and images under each person label in a dataset is crucial for the person Re-ID task. Therefore, we investigate the impact of different scales of image quantity and person ID quantity on the pre-trained backbone. Specifically, we involve various percentages of \textit{Diff-Person} in pre-training and then evaluate the fine-tuning performance on the target datasets. The results shown in Tab.~\ref{tab:imagescale} demonstrate that as the data scale increases, the performance of the pre-trained backbone gradually improves. It is mainly because our generation approaches can bring more diversity in attributes, which is a significant advantage compared to captured datasets. However, the performance tends to saturate when the data scale is large enough. This is mainly due to the model's capacity and ability limitations.

\begin{wrapfigure}{R}{0.6\columnwidth}
    \vspace{-0.15in}
    \centering
    \includegraphics[width=0.6\textwidth]{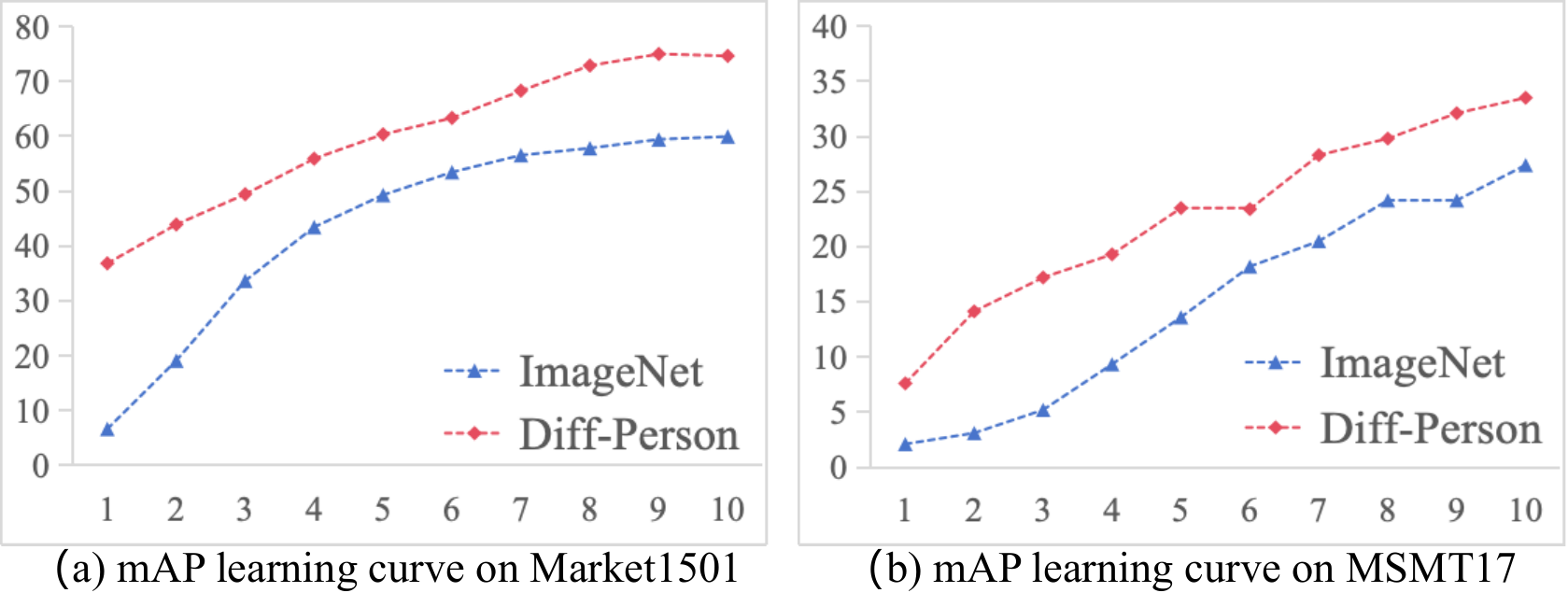}
    \vspace{-0.2in}
    \caption{Evaluation curves of mAP with different pre-training datasets.}
    \label{fig:map}
    \vskip -0.15in
\end{wrapfigure}
\textbf{Better Initialization and Faster Convergence}. 
Drawing many experimental validations, we rethink the factors contributing to the overall performance enhancement of the backbone pre-trained on \textit{Diff-Person} compared to the ImageNet-1K pre-trained backbone. The primary rationale behind this improvement is that \textit{Diff-Person} facilitates superior initialization and faster convergence.
We conduct a comparative analysis between our backbone and the ImageNet-1K pre-trained backbone during the early fine-tuning stage. On MSMT17 and Market-1501, our backbone attains superior mAP in the first epoch and consistently outperforms the ImageNet-1K pre-trained backbone throughout the first ten epochs. The mAP curves of fine-tuning MGN with our pre-trained backbone and the ImageNet-1K pre-trained backbone are illustrated in Fig.~\ref{fig:map}. This superiority is primarily attributed to our pre-trained backbone being exclusively based on person images, effectively eliminating the domain gap between ImageNet-1K and person image data. This alignment proves advantageous for the model in acquiring enhanced features tailored for person Re-ID.

\noindent \textbf{Performance on Transformer-based Backbone}.
In the main text, we use ResNet-50 as the backbone and fairly compare those methods. It is primarily attributed to the prevalent practice of person Re-ID, where most methods utilize ImageNet-1K pre-trained ResNet-50 as the backbone. In order to further demonstrate the generalization of our approach, we also conduct experiments with the Swin-Transformer backbone. We pre-train three different sizes of Swin-Transformer backbone based on \textit{Diff-Person} and make a fair comparison with the widely used Swin-Transformer-based method SOLIDER~\cite{chen2023beyond}. As shown in Tab.~\ref{tab:swin}, on Market-1501, we achieve performance gains of 2.5\%/0.7\% in mAP/Rank-1 with Swin Tiny; in MSMT17, we get performance gains of 4.4\% in mAP and 1.8\% in Rank-1 using the Swin Tiny model. These results demonstrate the superiority of our backbone. 
\begin{table*}[t]
        \caption{Comparing SOLIDER and \textit{Diff-Person} pre-trained Swin-Transformer backbone in Market-1501 and MSMT17.} 
    \centering
    \scalebox{0.8}{

    \begin{tabular}{ccccccc}
    \toprule    
        \multirow{2}{*}{Method} & \multicolumn{3}{c}{Market-1501}& \multicolumn{3}{c}{MSMT17}\\
         \cmidrule{2-7} 
         & Swin Tiny & Swin Small & Swin Base& Swin Tiny & Swin Small & Swin Base \\ \midrule
        SOLIDER~\cite{chen2023beyond} & 91.6/96.1 & 93.3/96.6 & 93.9/96.9  & 67.4/85.9 & 76.9/90.8 & 77.1/90.7  \\ \midrule
         \multirow{2}{*}{Diff-Person} &\textbf{94.1/96.8} & \textbf{94.7/97.0} & \textbf{95.0/97.1} &\textbf{71.8/87.7} & \textbf{77.2/91.0} & \textbf{77.3/91.2}\\
             & \textcolor{teal}{+2.5/+0.7} & \textcolor{teal}{+1.4/+0.4}& \textcolor{teal}{+1.1/+0.2}& \textcolor{teal}{+4.4/+1.8} & \textcolor{teal}{+0.3/+0.2}& \textcolor{teal}{+0.2/+0.5} \\  \bottomrule
    \end{tabular}}
    \label{tab:swin}
\end{table*}

\noindent \textbf{Comparisons with Self-supervised Methods}.
Another possible way that differs from our framework to provide good initialization is self-supervised pre-training. Both of them are ways to solve the problem of data scarcity. It is worth mentioning that they can co-exist and promote each other. For example, our synthesized images can further enlarge the scale of datasets for self-supervised learning to facilitate effective representation learning. To verify this, we pre-train SOLIDER using \textit{Diff-Person}. As shown in the Tab.~\ref{tab:self}, on Market-1501, we achieve a performance gain of 1.9\%/0.6\% in mAP/Rank-1 with Swin Tiny; In MSMT17, we get a performance gain of 1.8\%/0.9\% in mAP/Rank-1 with Swin Tiny. The results further illustrate the effectiveness of our generative strategy.
\begin{table*}[t]
    \centering
	\caption{Performance gain of training SOLIDER by using \textit{Diff-Person} dataset.}
    \scalebox{0.8}{

    \begin{tabular}{ccccccc}
    \toprule    
        \multirow{2}{*}{Method} & \multicolumn{3}{c}{Market-1501}& \multicolumn{3}{c}{MSMT17}\\ 
        \cmidrule{2-7}
         & Swin Tiny & Swin Small & Swin Base & Swin Tiny & Swin Small & Swin Base \\ \midrule
        SOLIDER~\cite{chen2023beyond} & 91.6/96.1 & 93.3/96.6 & 93.9/96.9 & 67.4/85.9 & 76.9/90.8 & 77.1/90.7  \\ \midrule
         \multirow{2}{*}{SOLIDER+Ours} &\textbf{93.5/96.7} & \textbf{94.0/97.0} & \textbf{94.4/97.2} &\textbf{69.2/86.8} & \textbf{77.4/91.1} & \textbf{77.4/90.9}\\
             & \textcolor{teal}{+1.9/+0.6} & \textcolor{teal}{+0.7/+0.4}& \textcolor{teal}{+0.5/+0.3} & \textcolor{teal}{+1.8/+0.9} & \textcolor{teal}{+0.5/+0.3}& \textcolor{teal}{+0.3/+0.2}\\  \bottomrule
       
    \end{tabular}}
    \label{tab:self}
\end{table*}

\textbf{Limitations}. Although a better pre-trained backbone is very beneficial for developing person Re-ID, our work still has some limitations. Mainly, we currently only generate more images based on widely used annotated datasets. To obtain a better pre-trained backbone, we can generate more data based on more annotated datasets in the future. At the same time, we can improve the weakly supervised datasets to expand the IDs with insufficient data and change the long-tail distribution.

\section{Conclusion}\label{sec6}
In this paper, we present a novel paradigm \textit{Diffusion-ReID} to efficiently augment and generate diverse images based on known identities without requiring any cost of data collection and annotation. We build an annotated person Re-ID dataset \textit{Diff-Person} based on a paradigm \textit{Diffusion-ReID} that generates a large number of images with ID consistency and attribute diversity. To our knowledge, this is the first person Re-ID dataset generated by diffusion models. We obtain a better pre-trained backbone and it outperforms the ImageNet-1K pre-trained backbone by a large margin.

\backmatter

\bmhead{Acknowledgements}
This work was supported by NSFC Project under Grant No.62176061. The authors would like to thank the anonymous reviewers for their valuable suggestions and constructive criticisms.

\section*{Declarations}
\begin{itemize}
\item Funding - This work was supported by NSFC Project under Grant No.62176061.
\item Conflict of interest/Competing interests - The authors declare that they have no competing interests
\item Ethics approval - Not applicable
\item Consent to participate - Not applicable
\item Consent for publication - Yes
\item Availability of data and materials - All used data is publicly available
\item Code availability - The code will be available after this paper is accepted.
\item Authors' contributions - Ke Niu, Haiyang Yu and Teng Fu mainly conducted experiments and wrote this manuscript. Xuelin Qian and Bin Li guided the method design and experiments. Xiangyang Xue put forward suggestions for method improvement. All authors read and approved this manuscript.
\end{itemize}

\bibliography{sn-bibliography}% common bib file

\end{document}